\documentclass[journal]{IEEEtran}
\usepackage{cite}
\usepackage{amsmath,amssymb,amsfonts}
\usepackage{algorithmic}
\usepackage{graphicx}
\usepackage{subcaption}
\usepackage{tikz}
\usepackage{tikz-3dplot} 
\usepackage{pgfplots}
\usepackage{hyperref}
\hypersetup{colorlinks,linkcolor={red},citecolor={blue},urlcolor={red}} 
\usepackage{xcolor}
\usepackage[table]{xcolor}
\usepackage{tabularx}
\definecolor{tableline}{RGB}{0,70,140} 
\definecolor{rowgray}{gray}{0.92}
\definecolor{headergray}{gray}{0.85}
\newcolumntype{C}{>{\centering\arraybackslash}X}

\ifCLASSINFOpdf
\else
\fi
\begin{document}
%
\title{Lightweight Spatiotemporal Highway Lane Detection via 3D-ResNet and PINet with ROI-Aware Attention}
%
%
%

\author{Sorna Shanmuga Raja and Abdelhafid Zenati.
        
\thanks{S. Raja \& A. Zenati are with Engineering Department, School of Science and Technology (SST), City, St George’s University of London, Northampton Square, London, EC1V 0HB, United Kingdom.
        \text{abdelhafid.zenati@city.ac.uk} 
        }
\thanks{}}

\maketitle

\begin{abstract}
This paper presents a lightweight, end-to-end highway lane detection architecture that jointly captures spatial and temporal information for robust performance in real-world driving scenarios. Building on the strengths of 3D convolutional neural networks and instance segmentation, we propose two models that integrate a 3D-ResNet encoder with a Point Instance Network (PINet) decoder. The first model enhances multi-scale feature representation using a Feature Pyramid Network (FPN) and Self-Attention mechanism to refine spatial dependencies. The second model introduces a Region of Interest (ROI) detection head to selectively focus on lane-relevant regions, thereby improving precision and reducing computational complexity.

Experiments conducted on the TuSimple dataset (highway driving scenarios) demonstrate that the proposed second model achieves 93.40\% accuracy while significantly reducing false negatives. Compared to existing 2D and 3D baselines, our approach achieves improved performance with fewer parameters and reduced latency. The architecture has been validated through offline training and real-time inference in the Autonomous Systems Laboratory at City, St George’s University of London. These results suggest that the proposed models are well-suited for integration into Advanced Driver Assistance Systems (ADAS), with potential scalability toward full Lane Assist Systems (LAS).
\end{abstract}


%
\IEEEpeerreviewmaketitle

\section{Introduction}
%
%
%
%
\IEEEPARstart{T}{he} 
Lane Departure Warning Systems (LDWS) represent a fundamental module within Advanced Driver Assistance Systems (ADAS), aiming to mitigate unintentional lane departures, which remain a major cause of severe road accidents \cite{lane1}. By utilising forward-facing cameras, LDWS continuously captures real-time road scenes and extracts lane boundary information through vision-based detection algorithms \cite{lane2}. The vehicle’s lateral displacement and heading angle relative to the lane centre are then estimated, enabling continuous monitoring of lane-keeping performance \cite{lane3}.

Contemporary LDWS implementations extend beyond simple boundary detection by incorporating road curvature estimation and vehicle dynamic states such as steering angle, yaw rate, and longitudinal velocity \cite{lane4}. This integration allows the system to predict imminent deviations based on the vehicle’s projected trajectory rather than solely reacting to boundary crossings. When an unintended departure is detected, typically in the absence of an active turn signal, the system delivers multimodal warnings including visual alerts, auditory signals, or haptic feedback such as steering wheel or seat vibrations \cite{tusimple2017dataset}. Such predictive and non-intrusive assistance enhances driver awareness while preserving human control authority, aligning with the broader objective of safe human–vehicle interaction in semi-autonomous driving contexts \cite{lane5}.

Subsequent advancements in deep learning have considerably improved LDWS performance, particularly in adverse conditions such as poor weather or occlusion \cite{lane6}. The integration of convolutional neural networks (CNNs) enables the system to learn robust spatial features, contributing to the development of more sophisticated Lane Assist Systems (LAS), which actively intervene through autonomous steering or braking~\cite{hou2019learning}. These lane monitoring technologies now operate alongside other ADAS functionalities such as adaptive cruise control and emergency braking~\cite{lin2017feature}, establishing LDWS and LAS as critical infrastructural elements for higher-level autonomous vehicles and forming a vital foundation for safer and more efficient transportation networks\cite{lane7}.

This paper proposes an enhanced deep learning model for lane detection that effectively integrates both spatial and temporal information. Two distinct networks were developed, each employing a 3D ResNet encoder~\cite{tran2015learning} and a Point Instance Network (PINet) decoder~\cite{ko2020keypoints}. The first network incorporates a Feature Pyramid Network (FPN)~\cite{lin2017feature} and a Self-Attention mechanism~\cite{vaswani2017attention} to capture multi-scale semantic representations and long-range spatial dependencies. The second network integrates a Region of Interest (ROI) detection head, which selectively refines features within relevant spatial areas for improved lane prediction and segmentation. The proposed second model achieved a classification accuracy of 93.40\% on the TuSimple dataset~\cite{tusimple2017dataset}, significantly reducing false negatives and outperforming prior 3D-CNN-based approaches in both accuracy and efficiency. To overcome the computational limitations of the standard 3D ResNet-50 architecture, a lightweight variant with shallower layers was developed to facilitate deployment in resource-constrained environments. Additionally, ablation experiments tested various feature extractors and decoding strategies to improve spatiotemporal feature learning. The final architecture is suitable for real-time lane detection tasks within Lane Warning Systems (LWS), with future potential for integration into Lane Assist Systems (LAS)~\cite{hou2019learning}. Offline and online experiments were conducted using real-world driving sequences at the Autonomous Systems Laboratory, City, St George’s University of London.

The remainder of the paper is organised as follows: Section \ref{sec2} reviews related work in traditional, 2D, and 3D lane detection methods. Section \ref{sec3} discusses the Theoretical background. Section \ref{sec4} details the proposed methodology, including network architectures, loss functions, and post-processing. Section \ref{sec5} describes the experimental setup, dataset, and evaluation metrics along with the results and comparative analysis, while Section \ref{sec6} concludes with final remarks and directions for future research.  

\section{Related Work}
\label{sec2}
Lane detection has emerged as a critical area within the development of Advanced Driver Assistance Systems (ADAS) and Autonomous Driving (AD) technologies. However, accurately identifying lane markings remains challenging due to variable real-world conditions such as occlusion, illumination changes, and worn or curved road markings. Early approaches relied on traditional computer vision techniques, such as the Hough Transform and Inverse Perspective Mapping (IPM), often supplemented with RANSAC for curve fitting. While computationally lightweight, these methods lacked robustness in dynamic or unstructured environments \cite{lane8}.

With the advent of deep learning, 2D Convolutional Neural Network (CNN) models significantly advanced lane detection performance. LaneNet~\cite{ko2020keypoints} introduced a pixel-wise segmentation framework using an encoder-decoder structure, while SCNN~\cite{pan2018spatial} extended this with spatial convolutions to preserve structural continuity. Models like Line-CNN~\cite{lin2020faster} and PolyLaneNet~\cite{liu2021rethinking} employed point- and parameter-based representations to model lane geometries more efficiently. Point Instance Network (PINet)~\cite{ko2020keypoints} treated lanes as clustered keypoints, enabling simultaneous multi-lane detection through instance segmentation. CLRNet~\cite{li2022clrnet} further enhanced performance via cross-layer refinement, fusing low and high-level features. However, all of these models process frames independently, neglecting temporal coherence.

To address this limitation, 3D deep learning architectures have emerged, enabling the joint modelling of spatial and temporal features. SegNet-ConvLSTM~\cite{lee2017vpgnet} combined convolutional segmentation with recurrent memory units to capture frame-to-frame dependencies. Similarly, a previous model integrating 3D-ResNet~\cite{tran2015learning} with PINet was proposed to process video sequences, utilising multiple output branches for confidence estimation and lane instance embedding. Despite improved robustness, this approach suffered from high computational demands and limited accuracy relative to advanced 2D models. The present work builds upon this foundation by introducing a lightweight 3D-ResNet encoder (2D+1D split) with ROI detection and attention mechanisms to enhance feature learning while improving inference efficiency.

\section{Theoretical Framework}
\label{sec3}
This section details the theoretical underpinnings and architectural design of the proposed lane detection system. The model leverages recent advancements in deep convolutional neural networks (CNNs), spatiotemporal learning, and attention-based mechanisms to robustly detect lane markings across varied driving scenarios. The architecture is composed of sequential modules: pre-processing, 3D-ResNet encoder, attention mechanism, Feature Pyramid Network (FPN) neck, Region of Interest (ROI) head, PINet-based decoder, and post-processing algorithms. Each component contributes to enhancing accuracy, efficiency, and robustness of detection.

\subsection{3D CNN Architecture Overview}

The end-to-end pipeline begins with pre-processing to normalize image sequences, correcting illumination and scale differences. Input data is fed into a 3D CNN encoder, which extracts both spatial and temporal features through convolutional layers operating on height, width, and time. These features are refined using self-attention, enhancing long-range dependencies, and further processed by an FPN neck to merge hierarchical features across multiple scales. A detection head generates preliminary predictions, and the decoder (PINet) reconstructs lane structures. Finally, post-processing, including RANSAC curve fitting and graph-based smoothing, ensures stable and continuous lane predictions.

\subsection{3D-ResNet Encoder}

The 3D-ResNet~\cite{tran2015learning} extends 2D residual networks by performing convolutions across three dimensions, thereby learning spatiotemporal patterns. This is crucial for lane detection, as temporal consistency across frames improves resilience against occlusion and noise. Each convolutional operation can be expressed as:
\begin{equation}
    y_{i,j,k} = \sum_{m,n,o} W_{m,n,o} \cdot X_{i+m, j+n, k+o} + b,
\end{equation}
where $W_{m,n,o}$ are kernel weights, $X$ is the input volume, and $b$ is bias.

Residual connections~\cite{he2016deep} prevent gradient vanishing and allow deep architectures to converge by bypassing information:
\begin{equation}
    y = f(x, \{W_i\}) + x.
\end{equation}
This enables efficient training of deeper models, critical for extracting complex patterns.

Additional components such as batch normalization~\cite{ioffe2015batch}, ReLU/LeakyReLU activations ~\cite{nair2010relu}, and dropout regularisation are employed to stabilise training, introduce non-linearity, and prevent overfitting, respectively.

\subsection{Attention Mechanism}

Traditional CNNs treat all features equally. Self-attention~\cite{vaswani2017attention} addresses this by assigning weights to input regions, focusing computation on lane-relevant features. The attention score is calculated as:
\begin{equation}
    \text{Self-Attention}(Q,K,V) = \text{softmax}\left(\frac{QK^T}{\sqrt{d_k}}\right)V,
\end{equation}
where $Q, K, V$ represent query, key, and value matrices derived from input features. This mechanism captures both local and global dependencies, crucial for detecting curved or occluded lanes.

\subsection{Feature Pyramid Network (FPN)}

The FPN~\cite{lin2017feature} integrates high-level semantic features with fine-grained spatial details through a top-down pathway and lateral connections:
\begin{equation}
    P_i = \text{Conv}(F_i) + \text{Upsample}(P_{i+1}).
\end{equation}
This hierarchical structure enhances multi-scale detection capability, allowing the network to capture lanes of varying widths, orientations, and visibility.

\subsection{PINet Decoder}

PINet~\cite{ko2020keypoints} serves as the decoder, specialising in point-based lane representation. It predicts discrete keypoints:
\begin{equation}
    P_k = \sigma(W_k * X + b_k),
\end{equation}
which are grouped using clustering algorithms based on Euclidean distance:
\begin{equation}
    D(p_1, p_2) = \sqrt{(x_1 - x_2)^2 + (y_1 - y_2)^2}.
\end{equation}
PINet employs multi-branch outputs: (i) a confidence map for lane points, (ii) offset maps to refine positions, and (iii) embeddings for lane instance separation. This ensures robust detection in complex environments like intersections.

\subsection{Loss Functions}

The network employs a multi-objective loss. Binary Cross-Entropy and Focal Loss~\cite{lin2017focal} handle class imbalance by emphasising difficult examples. Mean Squared Error ensures precise offset regression:
\begin{equation}
    \mathcal{L}_{\text{offset}} = \frac{1}{N} \sum_j ((x_j - \hat{x}_j)^2 + (y_j - \hat{y}_j)^2).
\end{equation}
Line Intersection-over-Union (LineIoU) refines lane alignment. The total loss is defined as:
\begin{equation}
    \mathcal{L}_{\text{total}} = \mathcal{L}_{\text{focal}} + \mathcal{L}_{\text{offset}} + \mathcal{L}_{\text{LineIoU}}.
\end{equation}

\subsection{Optimizer}

Adam~\cite{kingma2014adam} is employed for optimisation, dynamically adjusting learning rates using first and second moment estimates. This ensures faster convergence and stable training across deep architectures.

\subsection{Region of Interest (ROI)}

An ROI is applied to restrict attention to lane-relevant areas, reducing background noise and computational overhead:
\begin{equation}
    ROI(x,y) = I(x,y), \quad x \in [x_1,x_2], y \in [y_1,y_2].
\end{equation}
This focuses computation on road regions, improving detection robustness \cite{lane9}.

\subsection{Post-Processing}

RANSAC ~\cite{fischler1981ransac} curve fitting removes outliers and fits polynomial curves to lane points:
\begin{equation}
    f(x) = a_2x^2 + a_1x + a_0.
\end{equation}
Graph-based smoothing~\cite{liu2020graph} minimises abrupt changes in detected points by modelling them as weighted nodes and edges, with the shortest path computed via Dijkstra’s algorithm:
\begin{equation}
    P_{min} = \min \left( \sum_{i=1}^{n-1} w_{i,i+1} \right).
\end{equation}

\subsection{Evaluation Metrics}
Let TP, TN, FP, and FN denote true positives, true negatives, false positives, and false negatives, respectively. The performance is evaluated using standard classification metrics:
\begin{align}
\mathrm{Accuracy} &= \frac{TP + TN}{TP + TN + FP + FN}, \\
\mathrm{Precision} &= \frac{TP}{TP + FP}, \\
\mathrm{Recall} &= \frac{TP}{TP + FN}, \\
\mathrm{F1\text{-}Score} &= \frac{2 \, \mathrm{Precision} \cdot \mathrm{Recall}}
{\mathrm{Precision} + \mathrm{Recall}}.
\end{align}

These metrics capture both classification reliability and balance between false positives and false negatives.

Overall, this framework integrates spatiotemporal encoding, attention-based refinement, hierarchical feature fusion, and robust decoding, ensuring state-of-the-art lane detection performance under challenging real-world conditions.

\section{Methodology}
\label{sec4}

\subsection{Overview}
\begin{figure}[!h]
    \centering
    \includegraphics[width=\linewidth]{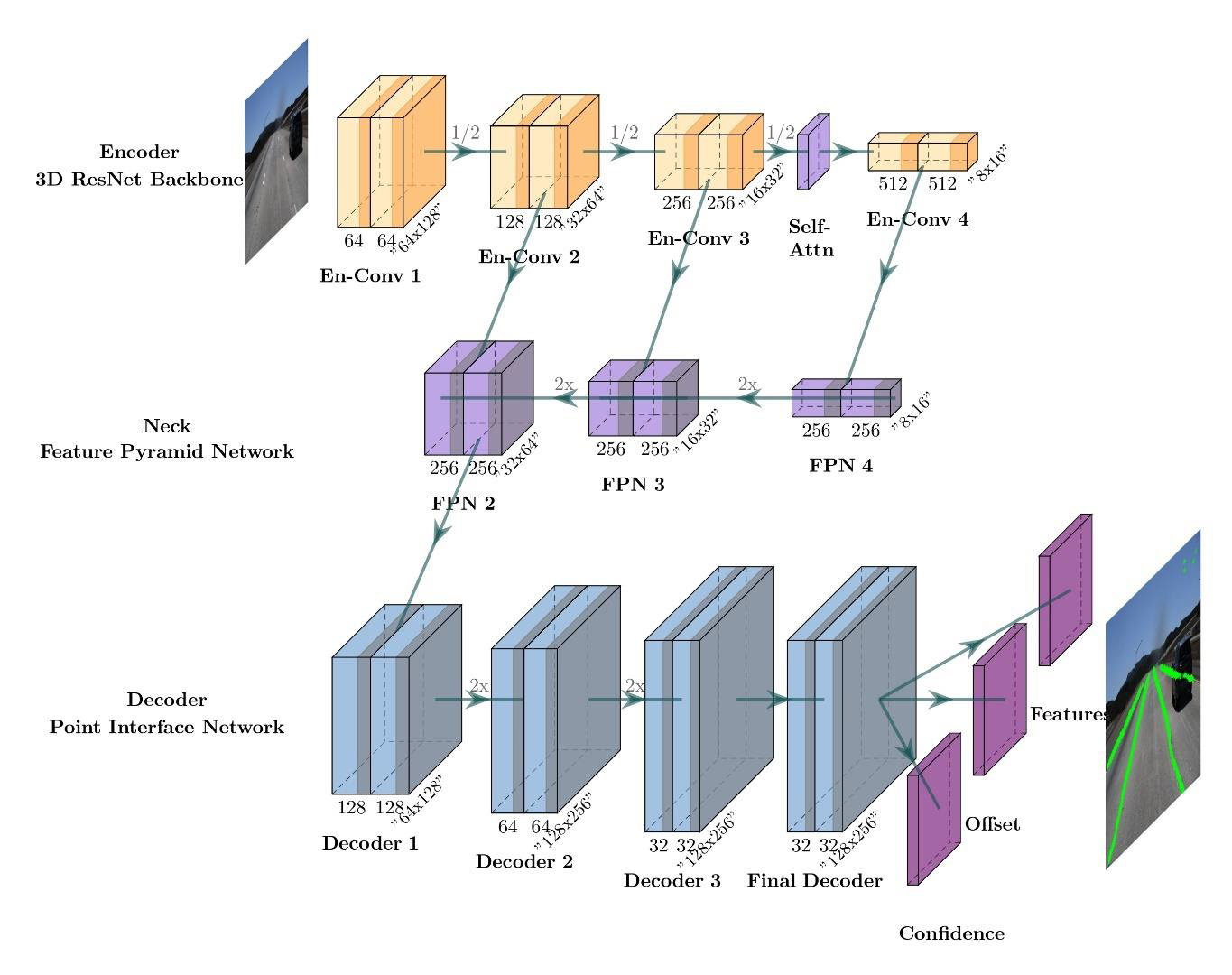}
    \caption{Proposed Network 1.}
    \label{fig:network1}
\end{figure}

\begin{figure}[!h]
    \centering
    \includegraphics[width=\linewidth]{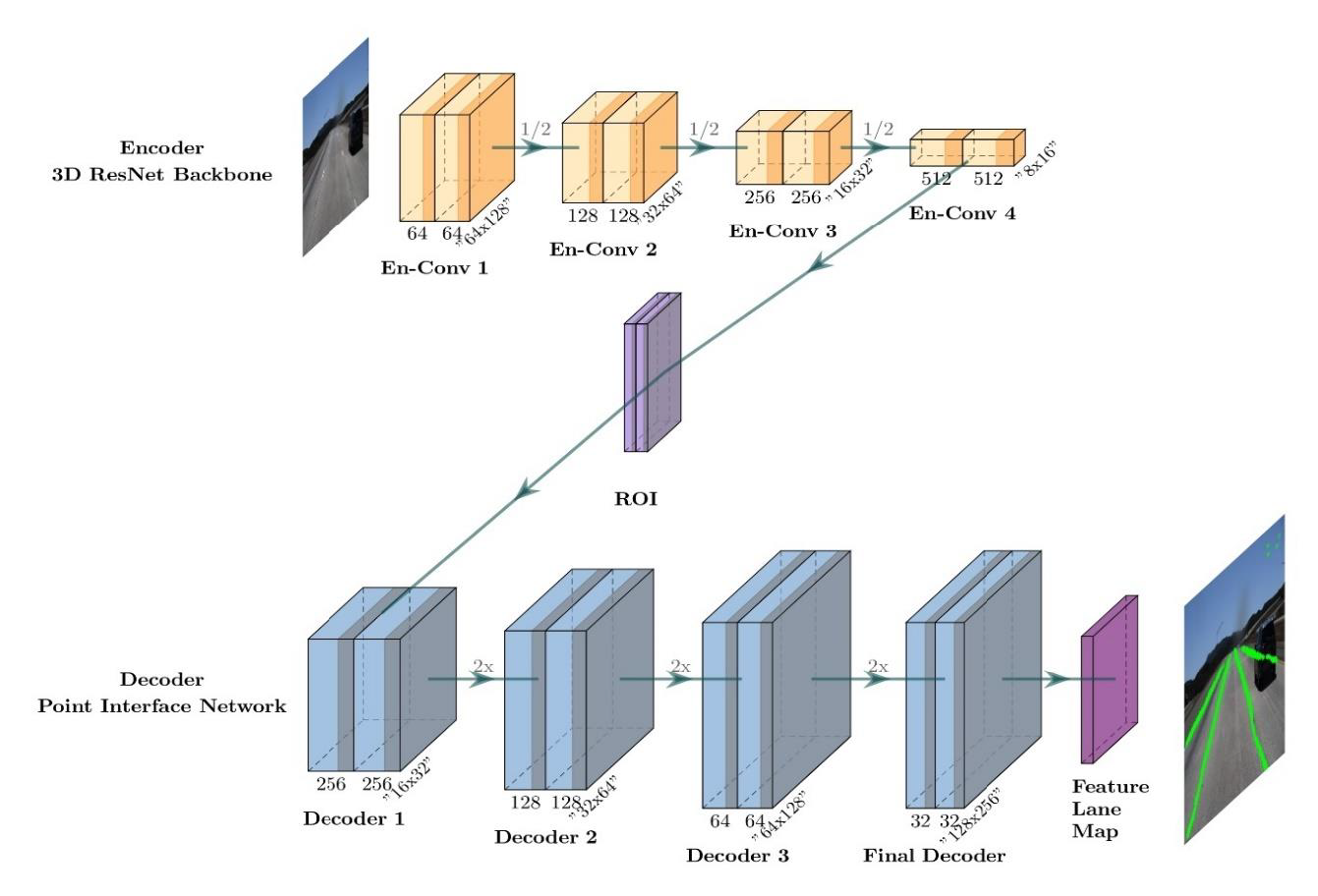}
    \caption{Proposed Network 2.}
    \label{fig:network2}
\end{figure}

The methodology of this study is centred on two proposed network architectures, both grounded in a shared 3D-ResNet encoder and PINet decoder baseline. Fig.~\ref{fig:network1} and~\ref{fig:network2} illustrate the design of these models. While both models incorporate common post-processing techniques such as RANSAC curve fitting and Graph-Based Smoothing to ensure geometrically coherent lane outputs, they differ significantly in their intermediate modules, which form the core of their novelty.

\subsubsection{Proposed Network 1}
Network 1 integrates a Self-Attention mechanism and Feature Pyramid Network (FPN) into the 3D-ResNet backbone. The 3D-ResNet encoder extracts spatiotemporal features, while self-attention enhances long-range dependencies, focusing on lane-relevant regions even under occlusion. The FPN aggregates multi-scale features, combining low-level spatial details with high-level semantics, improving robustness across varied driving conditions. The PINet decoder produces three outputs: confidence maps, offset predictions, and feature embeddings, trained jointly using a multi-branch loss to balance classification and regression.

\subsubsection{Proposed Network 2}
Network 2 introduces a Region of Interest (ROI) module in place of attention and FPN. The ROI restricts computation to lower image regions where lanes are concentrated, reducing false detections and improving computational efficiency. The decoder outputs a single feature map, optimised with a combined Focal Loss~\cite{lin2017focal} and Line IoU Loss, enhancing precision and ensuring alignment between predicted and ground-truth lanes.

\subsection{Dataset: TuSimple}
Experiments were conducted on the TuSimple dataset~\cite{tusimple2017dataset}, which provides 20-frame video clips depicting diverse highway driving conditions. It includes straight, curved, and broken lanes, under varying lighting and weather. Lane annotations are provided as discrete points in JSON format. Table~\ref{tab:tusimple} summarises its key features.

\begin{table}[!h]
\centering
\caption{Overview of TuSimple Dataset}
\label{tab:tusimple}
\rowcolors{2}{rowgray}{white}
\begin{tabularx}{\columnwidth}{l X}
\rowcolor{headergray}
\textbf{Attribute} & \textbf{Details} \\
Training Clips & 3,626 \\
Testing Clips & 2,782 \\
Frames per Clip & 20 \\
Resolution & $1280 \times 720$ pixels \\
Annotations & Lane point coordinates (JSON format) \\
Scenario & Multi-lane highway driving scenes \\
Conditions & Varied lighting and weather conditions \\
Evaluation Metrics & Accuracy, false positives (FP), false negatives (FN) \\
\end{tabularx}
\end{table}
\subsection{Hardware and Software Requirements}
Initial training and hyperparameter tuning were carried out on an NVIDIA GTX 1650 GPU (Dell G3 laptop). Final experiments, running up to 200 epochs, were performed on a High-Performance Computing (HPC) cluster. Implementation was in PyTorch, with supporting libraries including OpenCV for preprocessing, scikit-learn for evaluation, NetworkX for graph smoothing, and Matplotlib/Seaborn for visualisation. A dedicated Anaconda environment ensured dependency management.

\subsection{Pre-processing}
\begin{figure}[!h]
    \centering
    \begin{minipage}{0.32\linewidth}
        \centering
        \includegraphics[height=9cm,keepaspectratio]{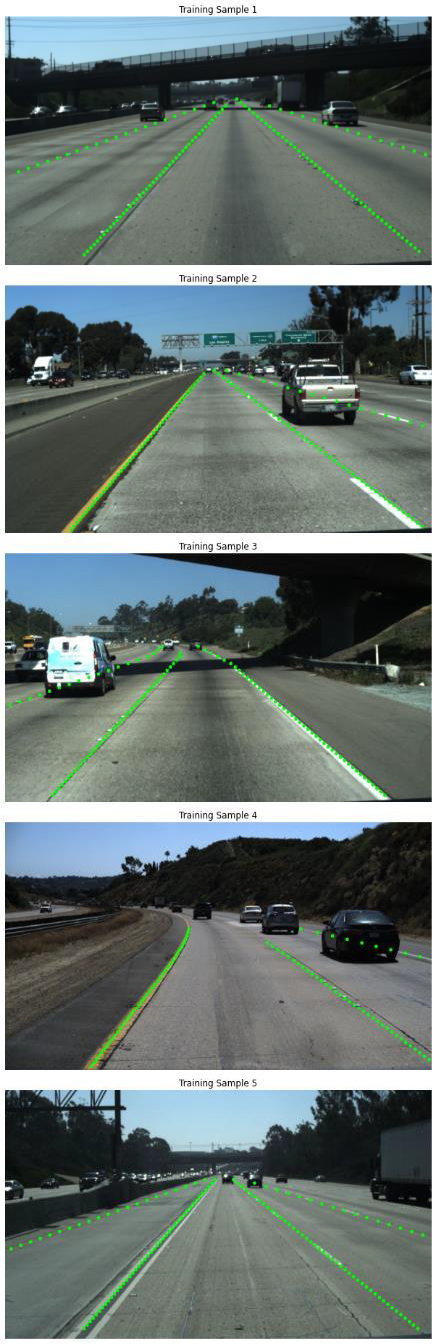}
        \caption*{(a)}
    \end{minipage}%
    \hfill
    \begin{minipage}{0.64\linewidth}
        \centering
        \includegraphics[height=9cm,keepaspectratio]{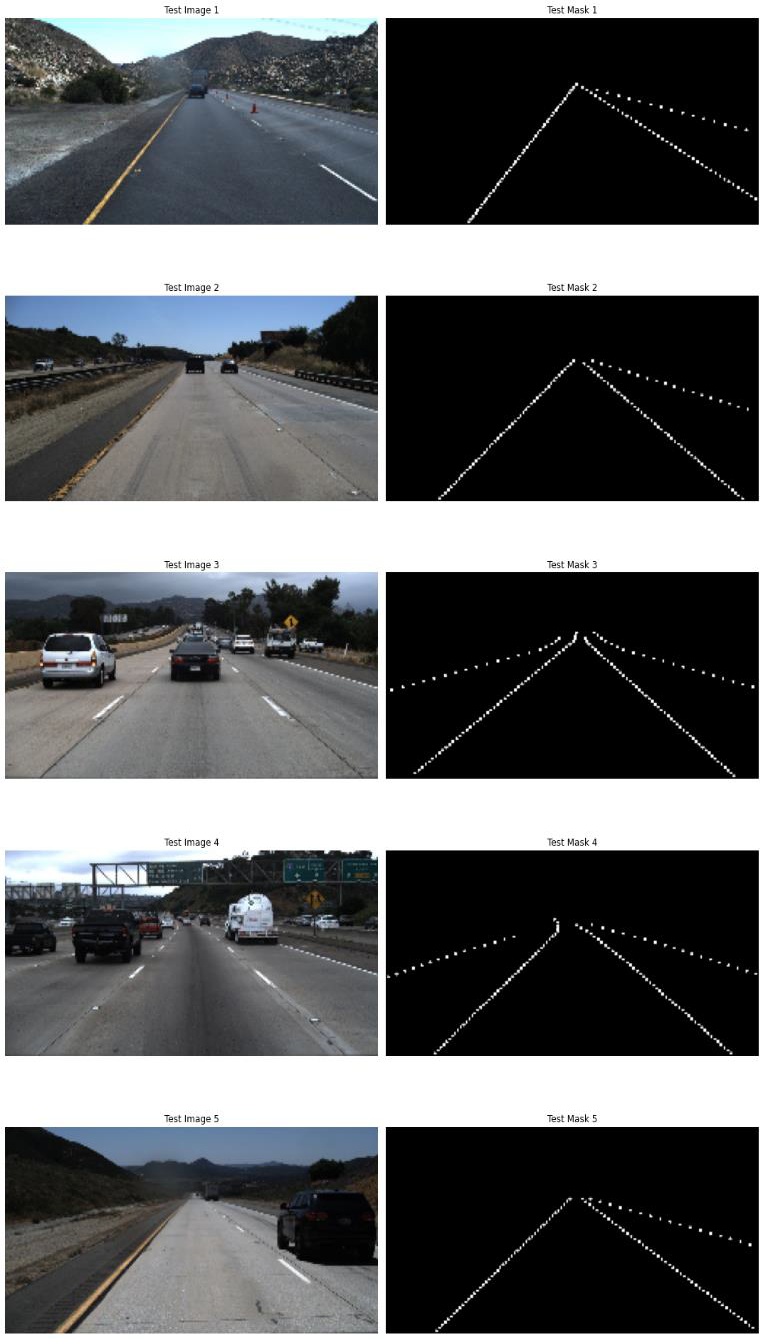}
        \caption*{(b)}
    \end{minipage}
    \caption{(a) Lane annotation overlay, (b) mask visualisation.}
    \label{fig:preprocess}
\end{figure}

Pre-processing involved annotation visualisation and data augmentation. Lane annotations were overlaid on images for inspection (Fig.~\ref{fig:preprocess}), and masks were generated to facilitate training. Augmentation included random flips, brightness adjustments, Gaussian noise, and rotations, improving generalisation under varied driving conditions.

\subsection{3D-ResNet Backbone}
The encoder employed a 3D-ResNet to capture temporal dependencies. Convolutional blocks used progressively increasing filters (64, 128, 256, 512). Self-attention (in Network 1) was applied after the third convolution to capture long-range feature dependencies, while dropout was used for regularisation. Network 2 omitted attention mechanism for computational efficiency.

\subsection{Self-Attention Implementation}

To enhance global contextual modelling, a self-attention module was inserted after the third convolutional block in Network 1. At this stage, feature maps retain sufficient spatial resolution while encoding higher-level semantic information, making it an appropriate location for long-range dependency modelling. The attention mechanism computes query, key, and value representations using $1 \times 1 \times 1$ convolutions, thereby preserving spatial dimensions while projecting features into an embedding space. Attention scores are obtained through scaled dot-product similarity between queries and keys, followed by softmax normalisation. The resulting attention weights are applied to the value representations to produce globally aggregated feature responses, which are subsequently fused with the original feature maps.

Table~\ref{tab:attention} provides a qualitative comparison between self-attention and commonly used alternative attention mechanisms.

\begin{table}[!h]
\centering
\caption{Comparison of Attention Mechanisms}
\label{tab:attention}
\rowcolors{2}{rowgray}{white}
\begin{tabularx}{\columnwidth}{l X X}
\rowcolor{headergray}
\textbf{Mechanism} & \textbf{Description} & \textbf{Limitation} \\
Spatial Attention & Emphasises informative spatial regions & Limited long-range dependency modelling \\
Channel Attention & Reweights feature channels & Does not explicitly model spatial relationships \\
Self-Attention & Models global feature interactions & Higher computational complexity \\
\end{tabularx}
\end{table}

\subsection{Feature Pyramid Network (FPN)}
The FPN aggregated multi-scale features by combining semantic-rich deep layers with high-resolution shallow layers. Feature maps were processed through lateral convolutions, top-down upsampling, and fusion, as summarised in Table~\ref{tab:fpn}.

\begin{table}[!h]
\centering
\caption{Progression of Feature Maps in FPN}
\label{tab:fpn}
\rowcolors{2}{rowgray}{white}
\begin{tabularx}{\columnwidth}{l l X}
\rowcolor{headergray}
\textbf{Layer} & \textbf{Resolution} & \textbf{Information Captured} \\
1st (shallow) & $128 \times 256$ & Fine-grained spatial structures and edge-level details \\
2nd & $64 \times 128$ & Mid-level texture and structural representations \\
3rd & $32 \times 64$ & Higher-order feature abstractions and pattern compositions \\
4th (deepest) & $16 \times 32$ & High-level semantic representations and contextual encoding \\
\end{tabularx}
\end{table}

\subsection{PINet Decoder}
The PINet decoder reconstructed lanes using transposed convolutions and upsampling. Multi-branch outputs included: (i) confidence maps, (ii) offset maps for refined keypoint localisation, and (iii) feature embeddings for lane instance separation. Outputs were trained with a combined focal and offset loss (Network 1) or focal and LineIoU loss (Network 2).

\begin{figure}[!h]
    \centering
    \begin{minipage}{0.66\linewidth}
        \centering
        \includegraphics[height=9cm,keepaspectratio]{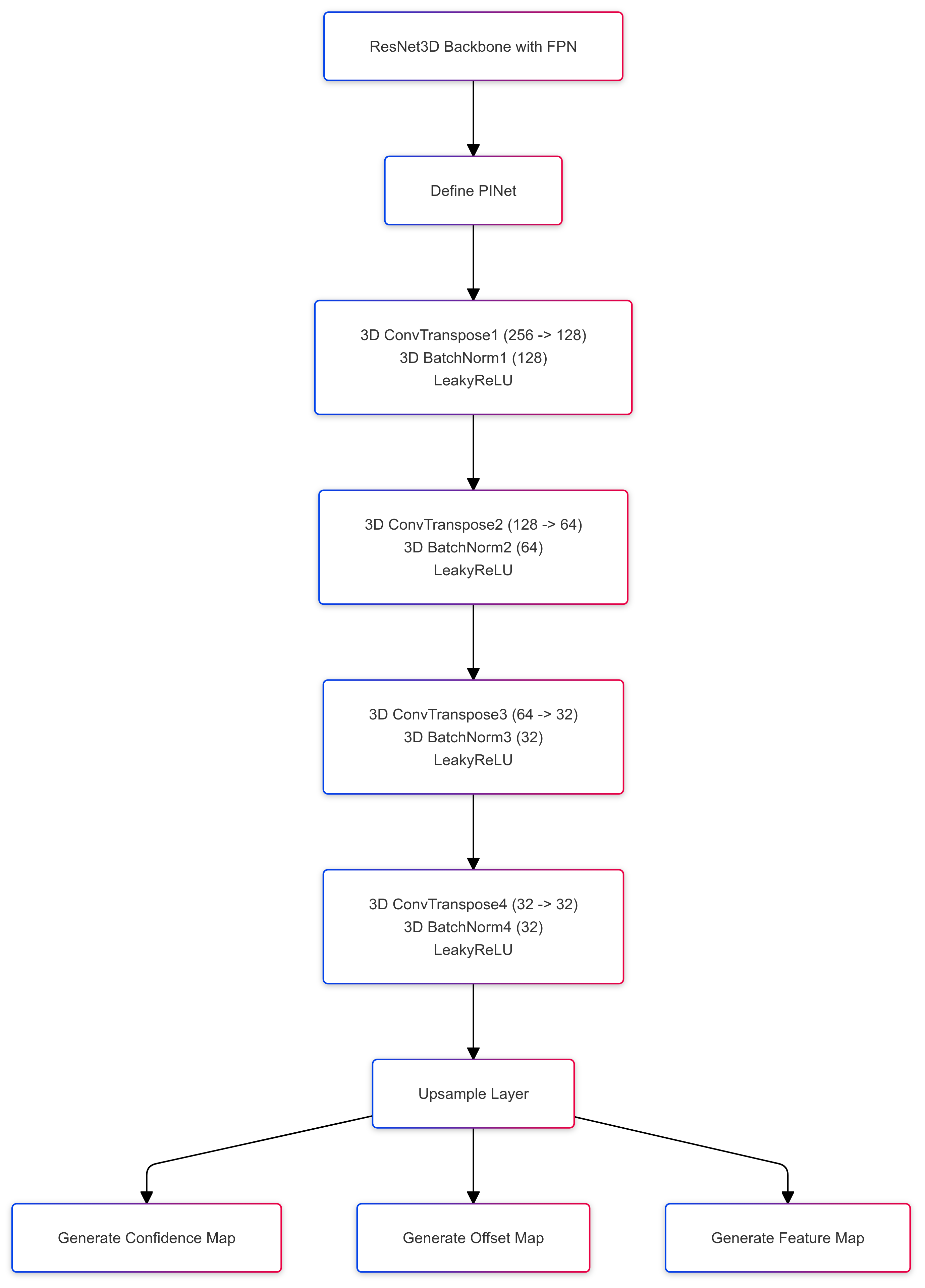}
        \caption*{(Network 1)}
    \end{minipage}%
    \hfill
    \begin{minipage}{0.3\linewidth}
        \centering
        \includegraphics[height=9cm,keepaspectratio]{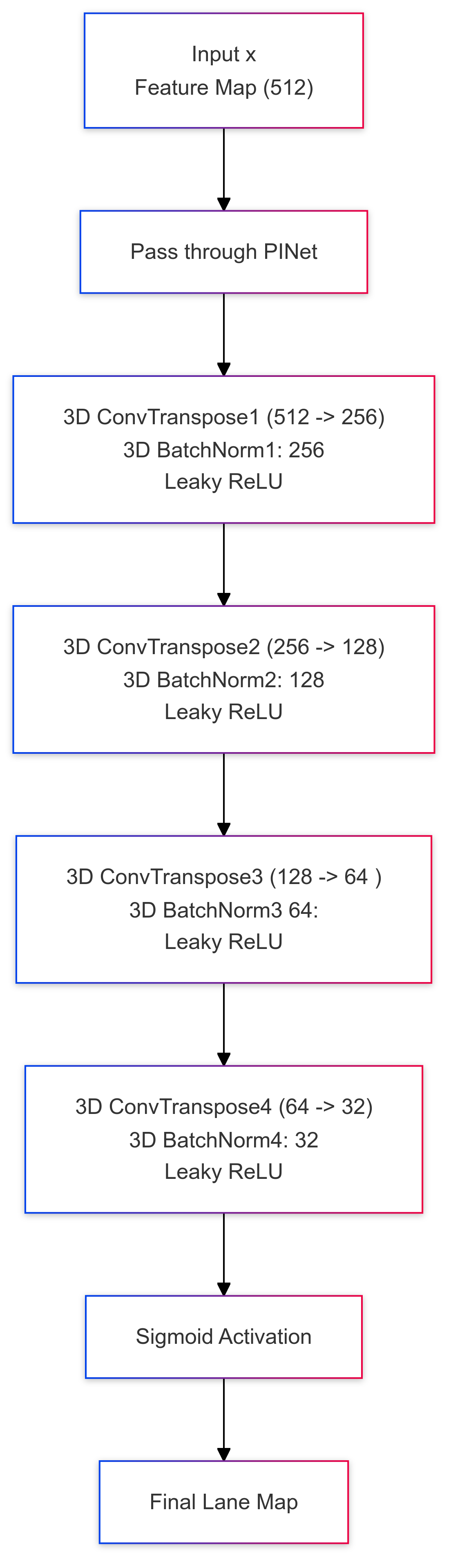}
        \caption*{(Network 2)}
    \end{minipage}
    \caption{PINet implementation with multi-branch outputs.}
    \label{fig:PINet}
\end{figure}

\subsection{Region of Interest (ROI)}
Network 2 employed ROI to restrict computation to lane-relevant image regions. This reduced background noise, improved efficiency, and enhanced alignment of lane predictions. Focal Loss emphasised hard-to-detect points, while LineIoU ensured geometric consistency.

\begin{figure}[!h]
    \centering
    \includegraphics[width=0.9\linewidth]{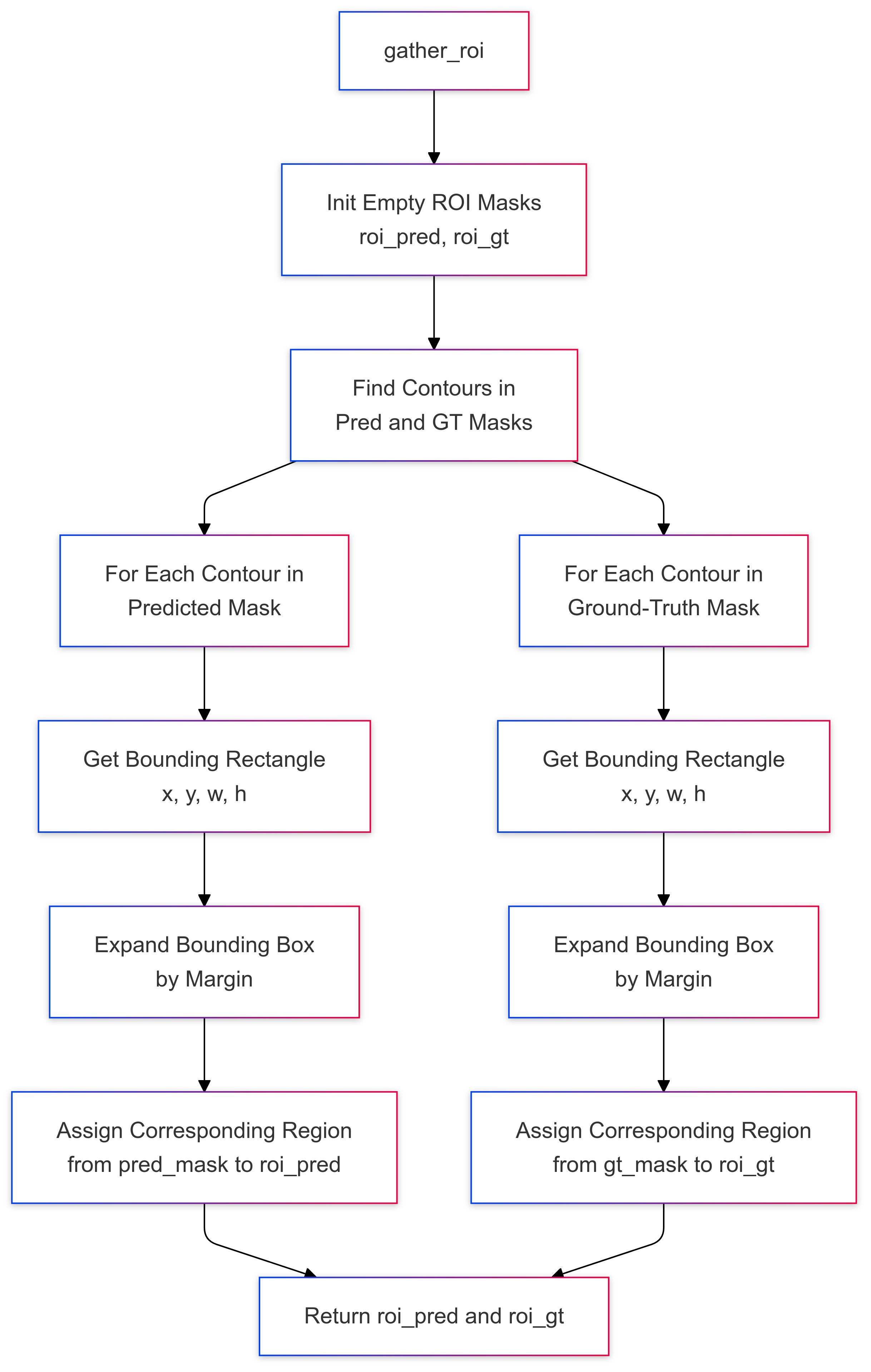}
        \caption{ROI implementation and integration with loss functions.}
    \label{fig:roi}
\end{figure}

\subsection{Post-Processing}
Post-processing included polynomial curve fitting using RANSAC and smoothing via graph-based algorithms~\cite{liu2020graph}. RANSAC removed outliers and generated stable fits, while graph smoothing minimised abrupt transitions, producing continuous and realistic lane lines.

\begin{figure}[!h]
    \centering
    \includegraphics[width=0.9\linewidth]{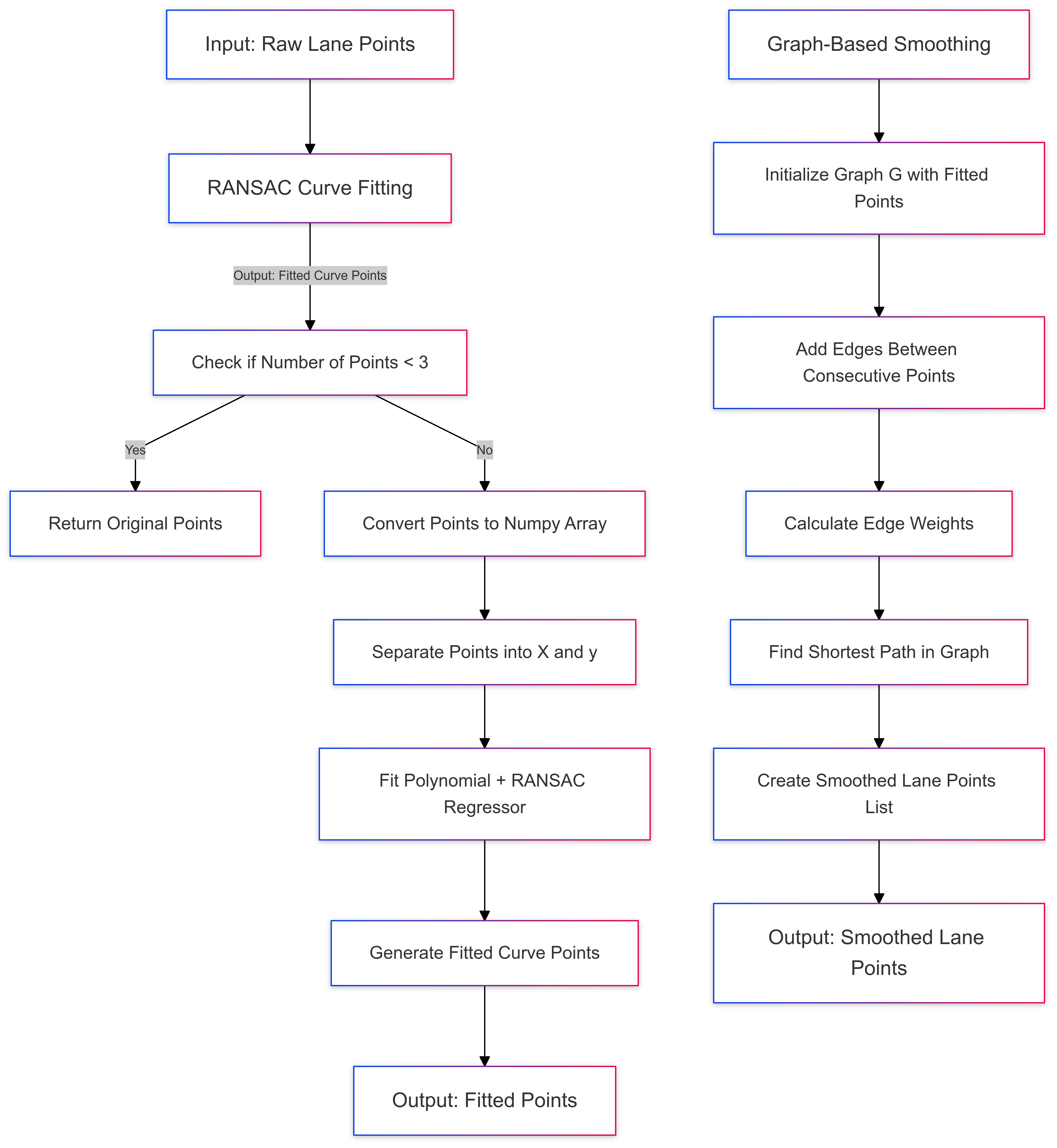}
    \caption{Post-processing pipeline with RANSAC and graph-based smoothing.}
    \label{fig:postprocessing}
\end{figure}

\section{Experimental Setup}
\label{sec5}

\subsection{Experimental Combinations}
Three experimental configurations were evaluated in parallel, each extending a common 3D-ResNet with PINet baseline and post-processing (RANSAC + graph smoothing). This allowed us to systematically assess the contribution of self-attention, FPN, ROI, and tailored loss functions. Table~\ref{tab:experiments} summarises the experimental setups.

\begin{table}[!h]
\centering
\caption{Summary of Experimental Configurations}
\label{tab:experiments}
\rowcolors{2}{rowgray}{white}
\begin{tabularx}{\columnwidth}{l l X}
\rowcolor{headergray}
\textbf{Exp.} & \textbf{Model} & \textbf{Enhancements} \\
1 & 3D-ResNet + PINet (Base Arch.) & Baseline configuration with RANSAC post-processing and graph-based smoothing \\
2 & + Self-Attention + FPN & Multi-scale feature fusion with multi-branch supervision \\
3 & + ROI + Focal + LineIoU & ROI-driven refinement with focal loss optimisation and improved lane alignment consistency \\
\end{tabularx}
\end{table}

\subsection{Experiment 1: Base Arc. (3D-ResNet + PINet)}
\begin{figure}[!h]
    \centering
    \begin{subfigure}{0.9\linewidth}
        \centering
        \includegraphics[width=\linewidth]{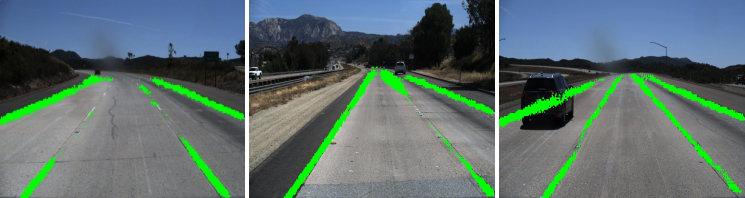}
        \caption*{(a) Predictions}
    \end{subfigure}
    \hfill
    \begin{subfigure}{\linewidth}
        \centering
        \includegraphics[width=\linewidth]{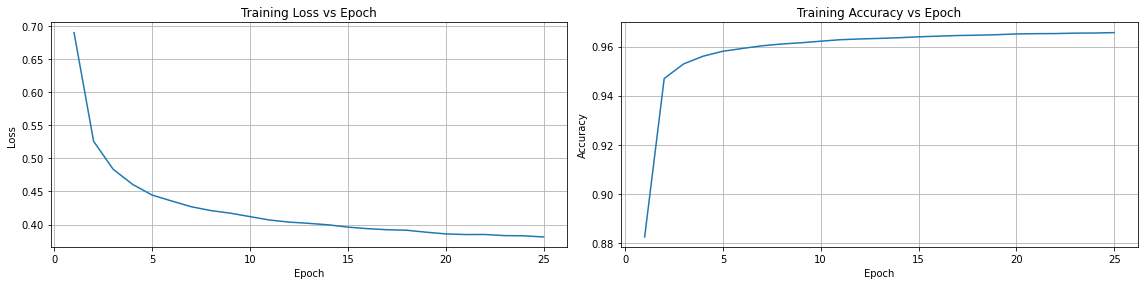}
        \caption*{(b) Training Curves}
    \end{subfigure}
    \hfill
    \begin{subfigure}{0.59\linewidth}
        \centering
        \includegraphics[width=\linewidth]{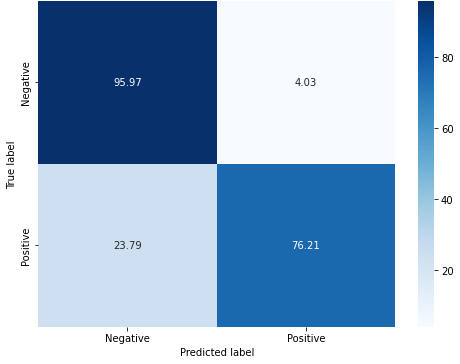}
        \caption*{(c) Confusion Matrix}
    \end{subfigure}
    \caption{Results of Experiment 1: lane predictions, training stability, and confusion matrix.}
    \label{fig:exp1_results}
\end{figure}

The Base Architecture achieved an accuracy of 86.09\% and precision of 94.98\%, but recall was limited (76.21\%), leading to an F1-score of 84.57\%. Predictions often broke down in curved or occluded lanes, confirming the need for enhanced feature fusion. Fig.~\ref{fig:exp1_results} shows sample predictions, training curves, and the confusion matrix.

\subsection{Experiment 2: Base Arc. + Self-Attention + FPN}
\begin{figure}[!h]
    \centering
    \begin{subfigure}{0.9\linewidth}
        \centering
        \includegraphics[width=\linewidth]{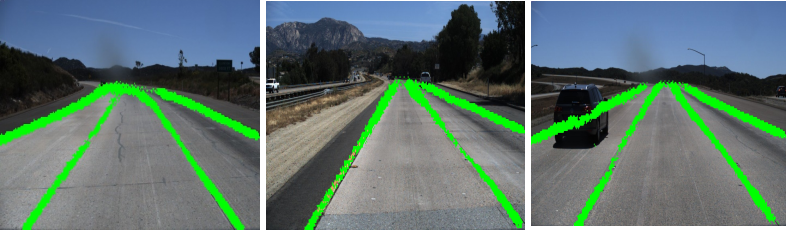}
        \caption*{(a) Predictions}
    \end{subfigure}
    \hfill
    \begin{subfigure}{\linewidth}
        \centering
        \includegraphics[width=\linewidth]{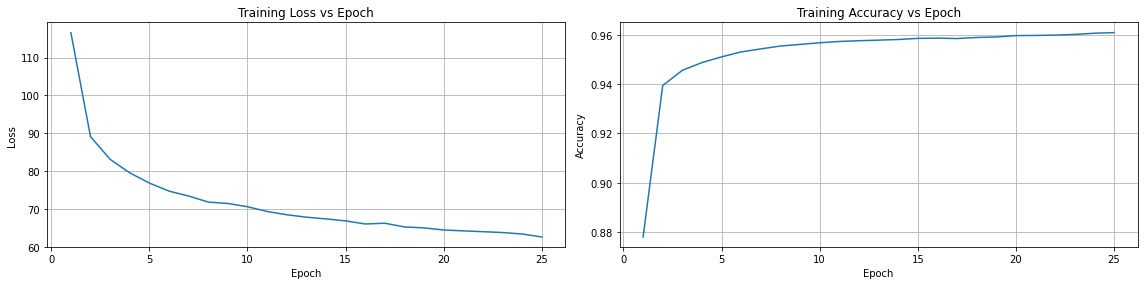}
        \caption*{(b) Training Curves}
    \end{subfigure}
    \hfill
    \begin{subfigure}{0.59\linewidth}
        \centering
        \includegraphics[width=\linewidth]{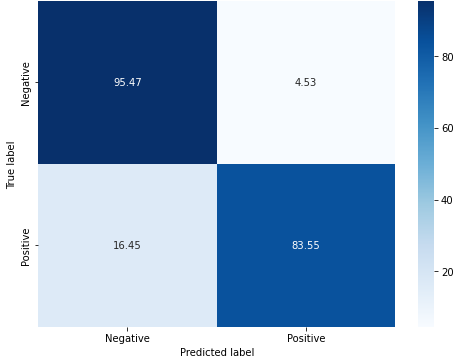}
        \caption*{(c) Confusion Matrix}
    \end{subfigure}
    \caption{Results of Experiment 2: lane predictions, training stability, and confusion matrix.}
    \label{fig:exp2_results}
\end{figure}

Adding self-attention and FPN improved robustness, especially under multi-lane or occluded conditions. Accuracy rose to 89.51\%, precision remained stable (94.86\%), and recall increased to 83.55\%, yielding an F1-score of 88.85\%. The network captured richer spatial dependencies but still faced difficulties in sharply curved or merging lanes. Results are illustrated in Fig.~\ref{fig:exp2_results}.

\subsection{Experiment 3: Base Arc. + ROI + Focal Loss + LineIoU}
\begin{figure}[!h]
    \centering
    \begin{subfigure}{0.9\linewidth}
        \centering
        \includegraphics[width=\linewidth]{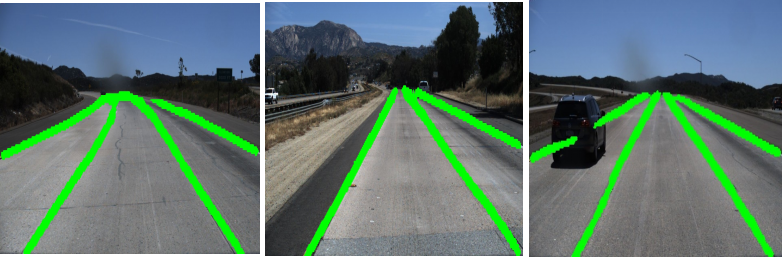}
        \caption*{(a) Predictions}
    \end{subfigure}
    \hfill
    \begin{subfigure}{\linewidth}
        \centering
        \includegraphics[width=\linewidth]{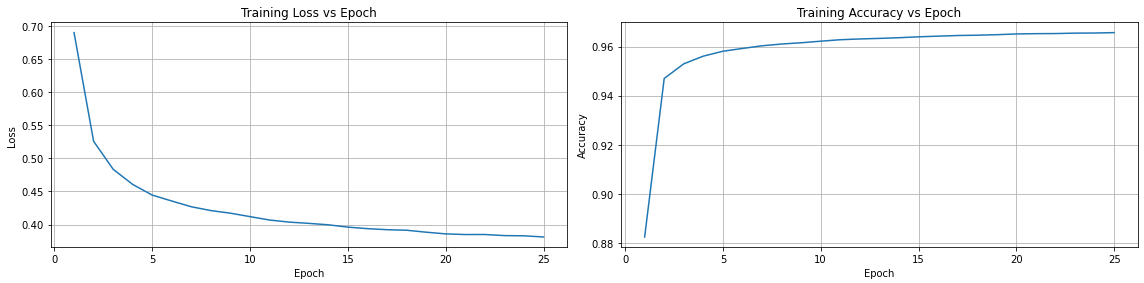}
        \caption*{(b) Training Curves}
    \end{subfigure}
    \hfill
    \begin{subfigure}{0.57\linewidth}
        \centering
        \includegraphics[width=\linewidth]{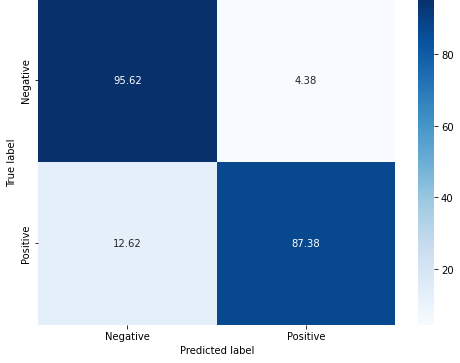}
        \caption*{(c) Confusion Matrix}
    \end{subfigure}
    \caption{Results of Experiment 3: lane predictions, training stability, and confusion matrix.}
    \label{fig:exp3_results}
\end{figure}

Experiment 3 achieved the best balance across metrics. Accuracy reached 91.50\%, with precision 95.23\%, recall 87.38\%, and F1-score 91.13\%. The ROI module reduced background noise and false positives, while Focal and LineIoU losses sharpened alignment with ground truth. Fig.~\ref{fig:exp3_results} highlights the improvements in prediction continuity.

\subsection{Extended Training (162 Epochs)}
\begin{figure}[!h]
    \centering
    \begin{subfigure}{0.9\linewidth}
        \centering
        \includegraphics[width=\linewidth]{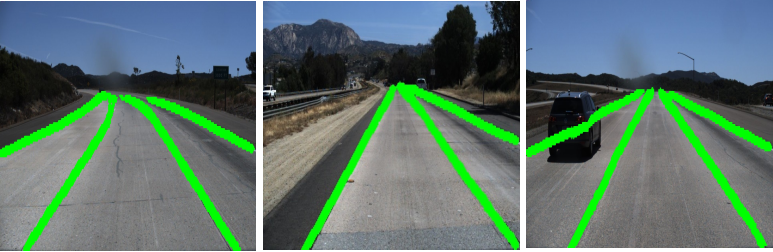}
        \caption*{(a) Predictions}
    \end{subfigure}
    \hfill
    \begin{subfigure}{\linewidth}
        \centering
        \includegraphics[width=\linewidth]{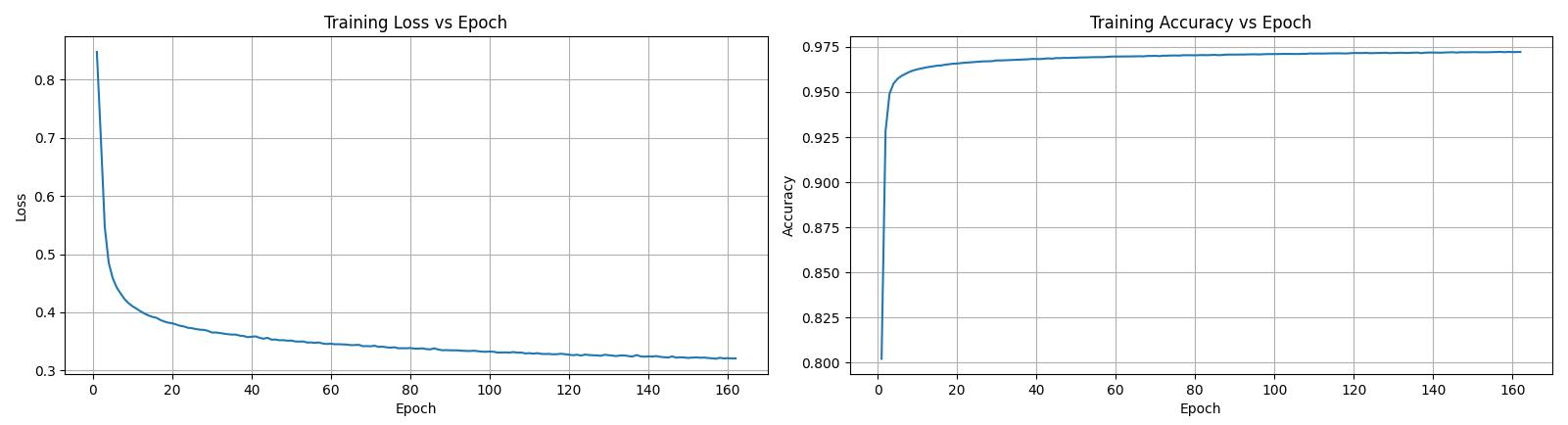}
        \caption*{(b) Training Curves}
    \end{subfigure}
    \hfill
    \begin{subfigure}{0.7\linewidth}
        \centering
        \includegraphics[width=\linewidth]{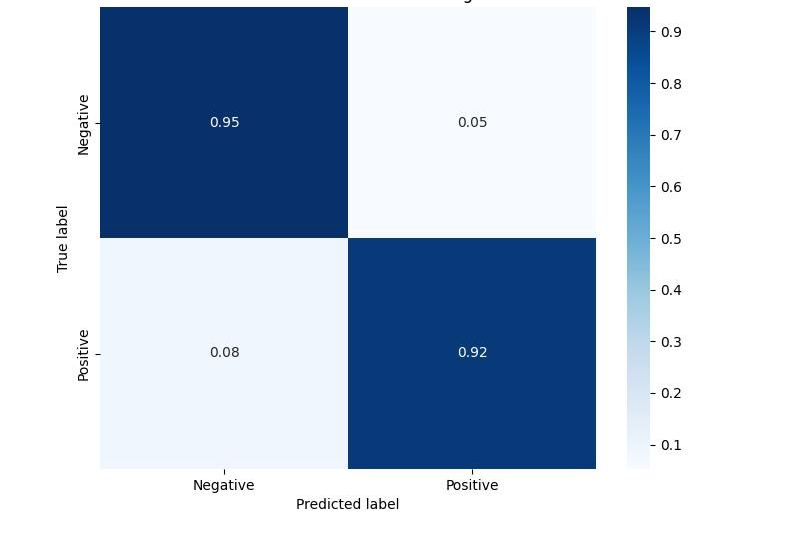}
        \caption*{(c) Confusion Matrix}
    \end{subfigure}
    \caption{Results of Experiment 4: lane predictions, training stability, and confusion matrix.}
    \label{fig:exp4_results}
\end{figure}

The best-performing model (Exp. 3) was trained for 162 epochs, reaching 93.33\% accuracy and 93.23\% F1-score. Predictions were sharper and more consistent in complex traffic and curved lanes, demonstrating readiness for ADAS deployment, Fig.~\ref{fig:exp4_results}.

\subsection{Ablation Study}
Alternative designs were tested (Table~\ref{tab:ablation}). While some achieved high precision (e.g., CABM and Line Prior, $\approx$96\%), their recall values ($<$35\%) indicated inability to detect complete lane structures. The Unet-based regression head performed worst (F1 = 31.65\%), confirming the superiority of our proposed architectures.

\begin{table}[!h]
\centering
\caption{Ablation Study Results}
\label{tab:ablation}

\small
\setlength{\tabcolsep}{3pt}        
\renewcommand{\arraystretch}{1.05} 

\rowcolors{2}{rowgray}{white}
\begin{tabularx}{\columnwidth}{>{\raggedright\arraybackslash}X c c c c}
\rowcolor{headergray}
\textbf{Model} & \textbf{Acc. (\%)} & \textbf{Prec. (\%)} & \textbf{Rec. (\%)} & \textbf{F1 (\%)} \\
Unet + Curve Fit & 55.36 & 67.50 & 20.67 & 31.65 \\
3D-ResNet + CABM & 64.56 & 96.07 & 30.35 & 46.13 \\
3D-ResNet + LSTM & 66.99 & 91.40 & 37.50 & 53.18 \\
3D-ResNet + Line Prior & 66.55 & 96.02 & 34.53 & 50.79 \\
\end{tabularx}
\end{table}

\section{Conclusion and Future Work}
\label{sec6}
This work presented a lightweight spatiotemporal highway lane detection framework using a 3D-ResNet encoder with PINet, enhanced by attention, multi-scale fusion, ROI, and tailored loss functions. Two variants were explored: one integrating Self-Attention and FPN, and another employing ROI with Focal and LineIoU losses. Experiments on the TuSimple dataset showed that the ROI-based model achieved the best results, reaching 93.33\% accuracy and reducing false negatives compared to both the baseline 3D-ResNet50 (91.34\%) and prior 3D-CNN approaches. These findings highlight the effectiveness of ROI-driven precision and advanced loss design in capturing complex lane geometries, making the proposed architecture well-suited for ADAS and scalable toward Lane Assist Systems.

\subsection{Future Work}
Future improvements can focus on three directions. (i) Hybrid encoder–decoder architectures combining convolutional and Transformer modules could better capture complex lane structures. (ii) Adaptive feature aggregation, such as integrating ROI after FPN, may improve robustness by dynamically weighting critical spatiotemporal cues. (iii) Dilated convolutions can be introduced to extend spatial context without added computation, enhancing long-distance lane detection. These enhancements would further strengthen accuracy and efficiency, enabling more context-aware lane detection for ADAS and autonomous driving.


%


\ifCLASSOPTIONcaptionsoff
  \newpage
\fi



%
\bibliographystyle{IEEEtran}
\bibliography{bibtex/bib/IEEEexample}

%




\end{document}